\documentclass[10pt,twocolumn,letterpaper]{article}

\usepackage[pagenumbers]{cvpr} % To force page numbers, e.g. for an arXiv version

\usepackage{pifont}
\usepackage{xcolor}
\usepackage{booktabs}
\usepackage{array}
\usepackage{tabularx}
\usepackage{multirow}
\usepackage{amsmath}
\usepackage{amssymb}

\definecolor{cvprblue}{rgb}{0.21,0.49,0.74}
\usepackage[pagebackref,breaklinks,colorlinks,allcolors=cvprblue]{hyperref}
\hypersetup{hidelinks}
\def\paperID{1053} % *** Enter the Paper ID here
\def\confName{CVPR}
\def\confYear{2026}

\title{Seeing Through the Noise: Improving Infrared Small Target Detection and Segmentation from Noise Suppression Perspective}

\author{
Maoxun Yuan$^{1}$ 
Duanni Meng$^{1}$ 
Ziteng Xi$^{1}$ 
Tianyi Zhao$^{1}$ 
Shiji Zhao$^{1}$ 
Yimian Dai$^{2}$ 
Xingxing Wei$^{1}$\\
$^{1}$Beihang University 
$^{2}$Nankai University\\
{\tt\small 
\{yuanmaoxun,mengduanni,xxwei\}@buaa.edu.cn,
yimian.dai@gmail.com
}
}
\begin{document}
\maketitle
\begin{abstract}
Infrared small target detection and segmentation (IRSTDS) is a critical yet challenging task in defense and civilian applications, owing to the dim, shapeless appearance of targets and severe background clutter. Recent CNN-based methods have achieved promising target perception results, but they only focus on enhancing feature representation to offset the impact of noise, which results in the increased false alarm problem. In this paper, through analyzing the problem from the frequency domain, we pioneer in improving performance from noise suppression perspective and propose a novel noise-suppression feature pyramid network (NS-FPN), which integrates a low-frequency guided feature purification (LFP) module and a spiral-aware feature sampling (SFS) module into the original FPN structure. The LFP module suppresses the noise features by purifying high-frequency components to achieve feature enhancement devoid of noise interference, while the SFS module further adopts spiral sampling to fuse target-relevant features in feature fusion process. Our NS-FPN is designed to be lightweight yet effective and can be easily plugged into existing IRSTDS frameworks. Extensive experiments on the IRSTD-1k and NUAA-SIRST datasets demonstrate that our method significantly reduces false alarms and achieves superior performance on IRSTDS task. The codes are available at \href{https://github.com/mengduann/NS-FPN}{https://github.com/mengduann/NS-FPN}.
\end{abstract}

\section{Introduction}
\label{sec:intro}

Infrared small target detection and segmentation (IRSTDS)  plays a critical role in various defense and civilian applications, including bird warning systems \cite{dai2021attentional}, sea rescue operations \cite{yuan2022translation,yuan2024c}, and aerial surveillance \cite{zhang2024e2e,yuan2024unirgb}. Due to the long imaging distance and lack of sufficient texture or structural information, infrared small targets (IRST) often appear as dim, shapeless spots, with extremely low signal-to-noise ratios (SNR) and signal-to-clutter ratios (SCR) \cite{dai2023one,yuan2024improving}. These characteristics severely hinder accurate detection, particularly in dynamic environments where targets are buried in heavy background clutter and exhibit weak thermal signatures. Therefore, developing robust and efficient IRSTDS methods that can accurately localize small targets under diverse and realistic infrared scenarios remains a challenging problem.

\begin{figure}[!t]
  \centering
  \includegraphics[width=0.8\linewidth]{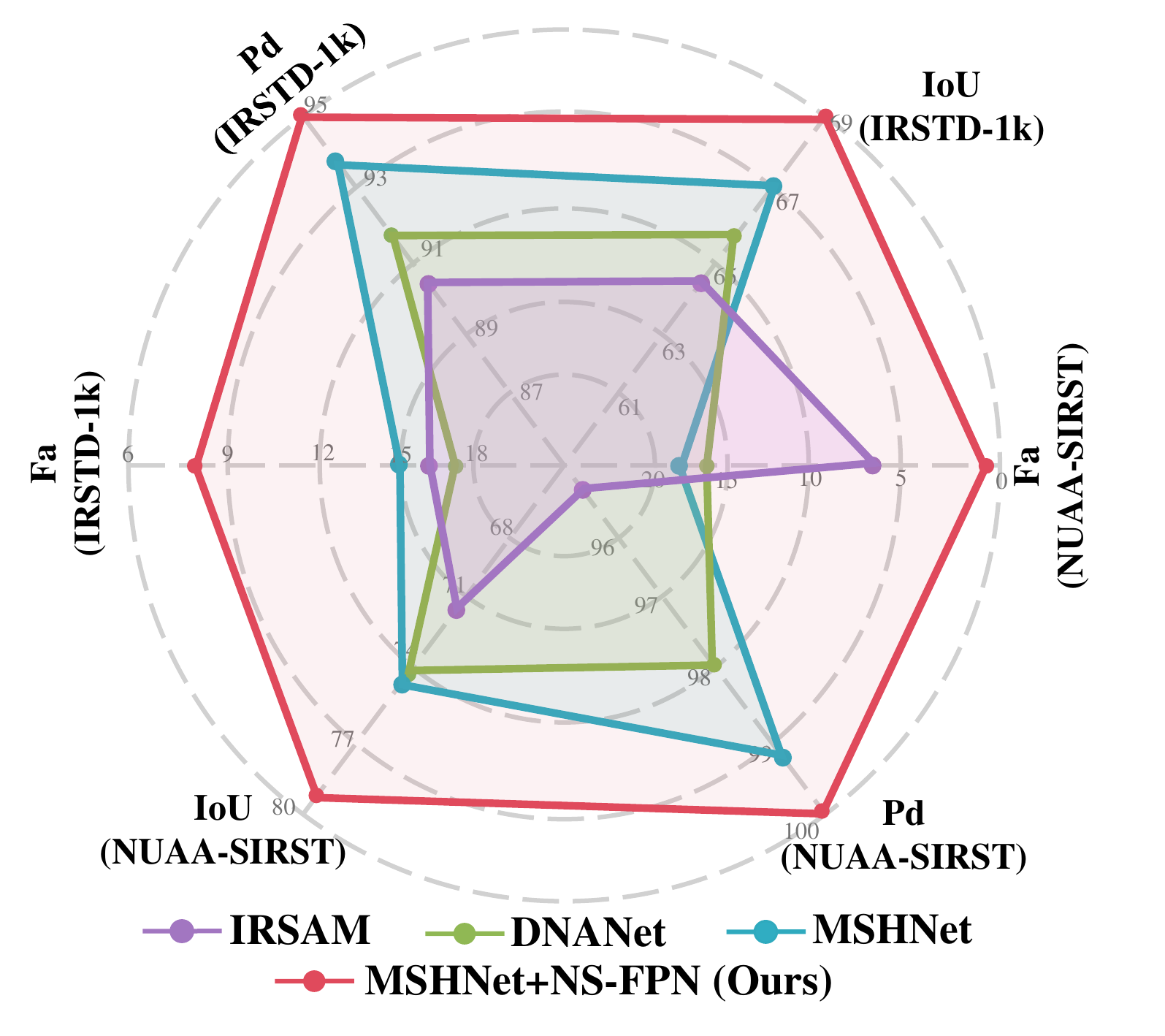}
  \vspace{-0.2cm}
  \caption{Performance comparison of our method with other methods on the IRSTD-1k and NUAA-SIRST datasets. The outer regions represent superior performance.}
    \vspace{-0.5cm}
  \label{fig:intro}

\end{figure}

Recent years have witnessed the focus of IRSTDS research shifting to CNN-based methods for detection \cite{dai2023one,zhang2023attention,li2023yolosr} and segmentation \cite{dai2021asymmetric,li2022dense,liu2024infrared,yuan2024sctransnet}, which concentrate on designing feature fusion structures to integrate high-level features with low-level details. For instance, DNANet \cite{li2022dense} is explored to mitigate deep information loss caused by pooling layers, and Liu \etal \cite{liu2024infrared} design a simple multi-scale head for the plain U-Net (MSHNet) to localize targets more precisely. In addition, to capture the essential structural features of small targets, IRPruneDet \cite{zhang2024irprunedet} proposes a wavelet channel pruning method, which believes that high-frequency components can discriminate the importance of features for IRSTDS. Recently, IRSAM \cite{zhang2024irsam} improves the Perona-Malik diffusion equation \cite{perona1994anisotropic} with a wavelet transform for image denoising and edge preservation, further improving detection performance. Although these methods have achieved satisfactory performance improvements by designing complex network structures, they primarily focus on enhancing the feature representation while neglecting the introduced false alarms, especially due to the emphasis on high-frequency components. As shown in Figure~\ref{fig:intro}, while the above methods exhibit satisfactory target localization performance in terms of $\mathrm{IoU}$ and $\mathrm{P_d}$, they concurrently exhibit high false alarm ($\mathrm{F_a}$) rates in IRSTDS task.

\begin{figure}[!t]
  \centering
  \includegraphics[width=1\linewidth]{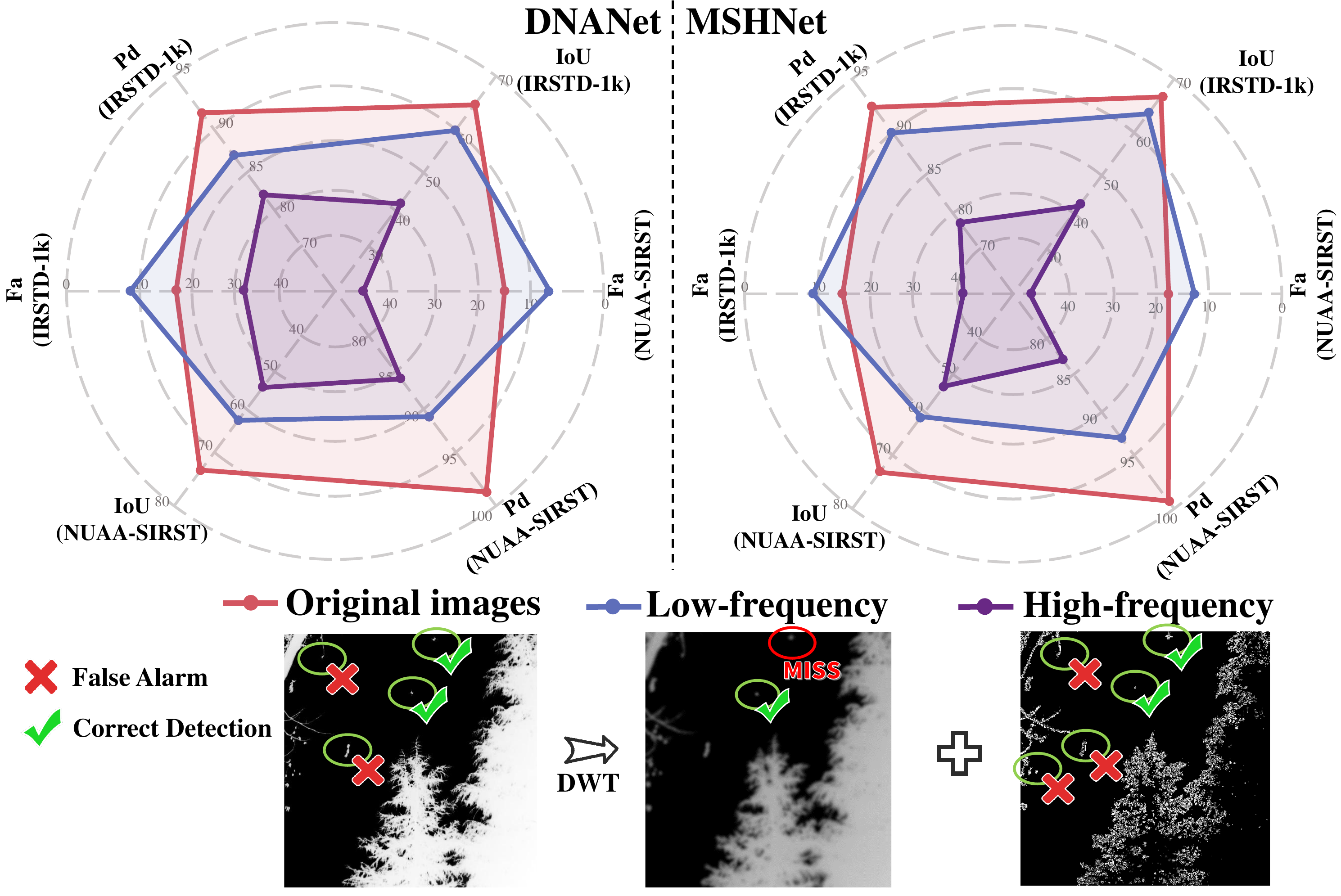}
  \caption{ The discrete haar wavelet transform (DWT) is utilized to decompose the original image into low and high frequency components. There is a crossover between the red and blue lines.}
   \vspace{-0.5cm}
  \label{fig:analysis}
\end{figure}

To solve this problem, we explore possible reasons for the increase in false alarms from the perspective of frequency domain. Since different frequency components have different functions in the image structure, high- and low-frequency components may play different roles in IRSTDS tasks. Thus, we decompose the original images as illustrated in Figure~\ref{fig:analysis} and observe the following characteristics:

\noindent\ding{202} \textbf{High-frequency components are crucial for IRSTDS tasks while also leading to an increase in false alarm rates.} Compared to low-frequency components, the $\mathrm{P_d}$ and $\mathrm{IoU}$ metrics of both methods show excellent performance on the original images, which indicates the image details (high-frequency components)  lost in low-frequency components are crucial for improving target localization. However, these high-frequency components also contain noise interference, leading to an increase in false alarm rates.

\noindent\ding{203} \textbf{Low-frequency components will degrade target localization performance but can also serve as valuable cues to reduce false alarms.} As shown in Figure~\ref{fig:analysis}, while the performance on original images surpasses that on low-frequency components, the $\mathrm{F_a}$ metric reveals that low-frequency components achieve the best false alarm suppression performance. This superior performance highlights the potential of low-frequency components as effective features for suppressing noise in high-frequency components.

Inspired by the above, we pioneer in suppressing the noise in feature fusion to ensure improving target localization performance while reducing false alarm rates. Therefore, we firstly propose a \textbf{l}ow-frequency guided \textbf{f}eature \textbf{p}urification module (LFP) to suppress the noise features by purifying high-frequency components. Specifically, LFP begins with a 2D discrete wavelet transform (DWT) to decompose features into the frequency domain, then utilizes low-frequency features to predict the response at potential target locations. This prediction serves as a weighted map to refine high-frequency features, followed by gated gaussian filtering to further suppress less confident features. Finally, inverse DWT transformation yields noise-suppressed features while preserving high-frequency enhancement. Furthermore, to avoid interference from surrounding background noise, we design a \textbf{s}piral-aware \textbf{f}eature \textbf{s}ampling module (SFS) that performs spiral sampling in feature fusion. Specifically, SFS employs dynamic sampling based on the intensity distribution characteristics (spiral shape) and calculates similarity to acquire the target-relevant features, further mitigating the impact of noise interference. 

Different from previous methods that focus on designing complex network structures, our paper aims to propose a lightweight yet effective way to improve performance. Thus, we integrate the LFP and SFS modules into the feature pyramid network (FPN) to construct a \textbf{n}oise-\textbf{s}uppression \textbf{FPN}, named \textbf{NS-FPN}. In this way, the LFP module can be used to replace the $1\times1$ convolutions in FPN to achieve feature enhancement devoid of noise interference, while the SFS module substitutes the upsampling operations to achieve structured sampling and fusion of target-relevant features. These efficient designs allow our NS-FPN to be easily plugged into existing IRSTDS frameworks. Figure~\ref{fig:intro} shows that our NS-FPN can effectively reduce false alarms and achieve superior target localization performance. In summary, our contributions in this paper are highlighted as follows:

\ding{202} We reveal the increased false alarm rate faced by current CNN-based IRSTDS methods from a frequency domain and pioneer in improving detection and segmentation performance from noise suppression perspective.

\ding{203} We propose a novel feature pyramid network (NS-FPN) that suppresses noise interference while enhancing target features using a low-frequency guided feature purification module and a spiral-aware feature sampling module.

\ding{204} Extensive experiments on the public IRSTDS datasets demonstrate that our NS-FPN can effectively reduce false alarms to further improve detection performance.

\section{Related Work}

\subsection{IRST Detection and Segmentation Networks}
With the advancement of deep learning techniques, CNN-based methods have emerged as the dominant paradigm, enabling automatic multi-layered feature learning for IRSTDS. For CNN-based detection, Dai \etal ~\cite{dai2023one} proposed a one-stage cascade refinement network (OSCAR) that uses the results of high-level heads to modulate the predictions of low-level heads, completing the detection of a target from coarse to fine, and Yang \etal ~\cite{yang2024eflnet} introduced a dynamic head mechanism to adaptively adjust feature responses to different spaces and channels, thus improving the ability of network to focus on tiny targets.
As for CNN-based segmentation, ISNet~\cite{zhang2022isnet} improved the accuracy of target shape prediction by introducing an edge reconstruction mechanism, while DNANet~\cite{li2022dense} maintained the information of small targets in deep layers through dense interactions between features from the same and different layers. Recently, Liu \etal ~\cite{liu2024infrared} used a novel scale and location sensitive (SLS) loss function to correct multi-scale prediction results, which can further improve the detection performance of existing methods. Concurrently, IRPruneDet \cite{zhang2024irprunedet} proposed a wavelet structure-regularized soft channel pruning strategy to improve the network inference efficiency.

Instead of focusing on complex structure design, this paper achieves feature enhancement devoid of noise interference, thus effectively reduce false alarms to further improve detection and segmentation performance.

% Dai \etal \cite{dai2021asymmetric} proposed an asymmetric contextual modulation (ACM) to better embed advanced semantics into the network without overwhelming the details of small targets. 
% UIUNet \cite{wu2022uiu} realized multi-scale representation learning of image information and enhanced global and local feature differences by embedding a small U-Net into a large U-Net backbone.
% WTAPNet \cite{he2024wtapnet} is based on DWT and IDWT to improve the downsampling and feature fusion module to reduce the information loss of small targets. 

% \begin{figure}[t]
%   \centering
%   \includegraphics[width=\linewidth]{figure/NS-FPN-framework.pdf} 
%   \caption{The overall structure of our NS-FPN and its conponents. Note that, convolutions are omitted for better visualization and $Y_4$ does not contains SFS module. }
%   \label{fig:NS-FPN}
% \end{figure}

\begin{figure*}[t]
  \centering
  \includegraphics[width=\linewidth]{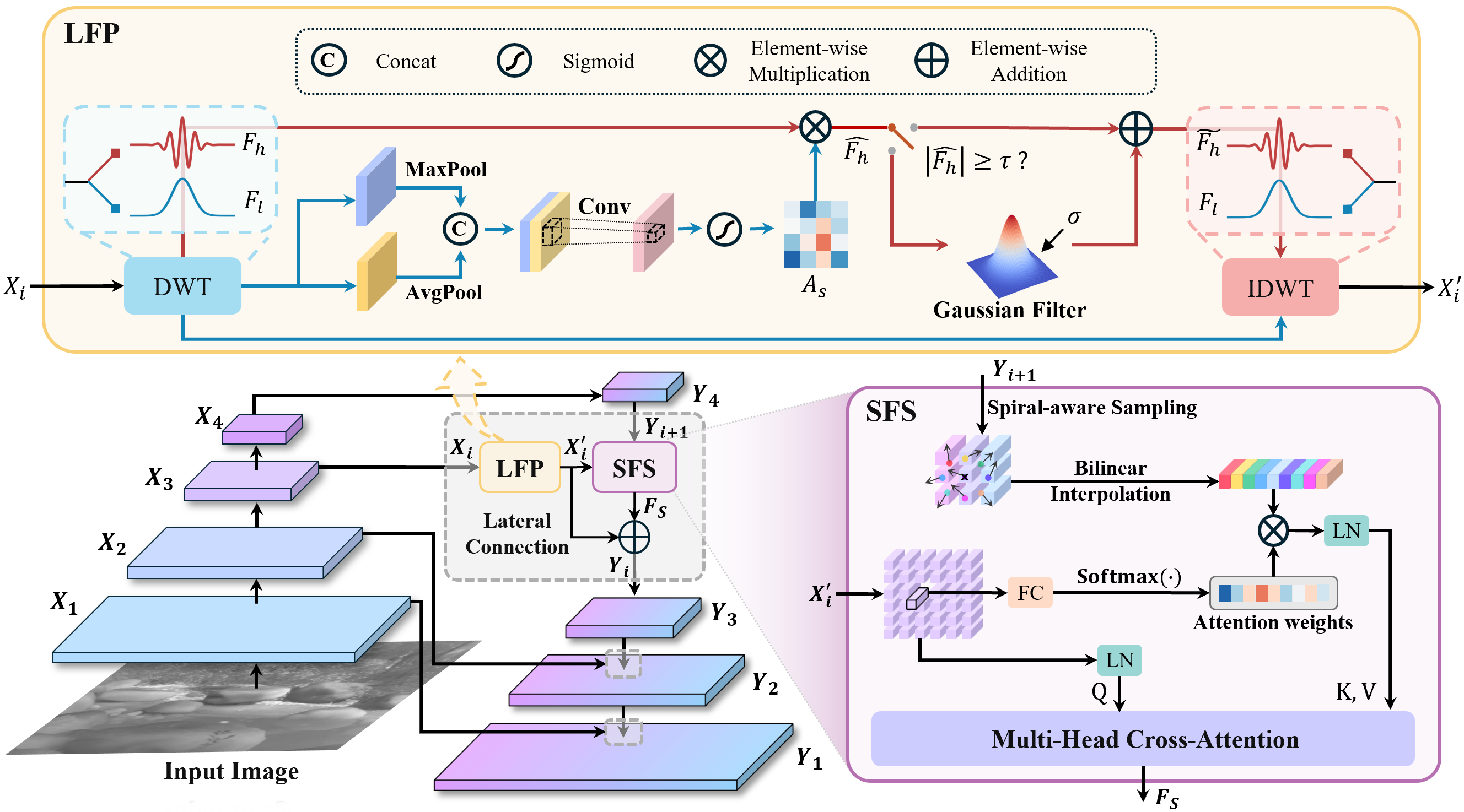} 
  \caption{The overall structure of our NS-FPN and the details of its conponents. For each scale, the feature is first fed into the LFP module, which generates a target-relevant spatial attention map based on low-frequency components to guide two-stage feature purification of high-frequency features. Subsequently, the purified feature is fused with the upper-layer feature through the SFS module to achieve semantic complementation according with spatial distribution prior of infrared small targets. Note that, convolutions are omitted for better visualization and $Y_4$ does not contain SFS module.}
  % \caption{The overall structure of our NS-FPN and its conponents. Note that, convolutions are omitted for better visualization and $Y_4$ does not contain SFS module. }
  \label{fig:NS-FPN}
\end{figure*}

% Overview of our proposed SEF-DETR. The input infrared image is processed through two complementary paths. The top
% branch shows the Frequency-guided Patch Screening (FPS) Module, which produces a pixel-wise target-relevant density map indicating
% potential target regions. This map is then employed at two critical stages: in the Dynamic Embedding Enhancement (DEE) Module to
% refine multi-scale embedding features, and in the Reliability-Consistency-aware Fusion (RCF) Module to guide the selection of initial
% queries. Finally, these refined queries are fed into the Transformer Decoder to perform accurate target localization and classification.

\subsection{Feature Pyramid Network}
Feature Pyramid Network (FPN)~\cite{lin2017feature} emerged as a fundamental structure for general object detection, which can integrate high-level features with low-level details through a top-down connection. However, its linear summation may not be the most effective method for feature fusion. Based on it, path aggregation network (PANet)~\cite{liu2018path} improved the FPN by introducing a bottom-up path enhancement, which facilitates the reuse of low-level features and shortens information flow paths. Besides, bidirectional feature pyramid network (BiFPN)~\cite{tan2020efficientdet} removed single-input nodes from PANet and introduced an adaptive weighted feature fusion mechanism to improve detection performance. In contrast to above FPNs that rely on manually crafted designs, NAS-FPN~\cite{ghiasi2019fpn} leveraged neural architecture search to automatically explore the topological space of feature fusion networks and discovered optimal connection strategies. Recently, 
HS-FPN \cite{shi2025hs} was designed for tiny object detection in visible images. However, it over-relied on high-frequency features determined by a preset spectral range and neglected the importance of low-frequency features, making it ineffective for IRST task. Unlike above methods, our work fully utilizes low-frequency components to adaptively guide high-frequency components, thereby actively suppressing noise in feature fusion process, which is a specific design for the IRSTDS.

\section{Method}
% \subsection{Noise-Suppression FPN (NS-FPN)}
In this section, we introduce our noise-suppression feature pyramid network (NS-FPN) and its components. As illustrated in Figure~\ref{fig:NS-FPN}, NS-FPN adopts a similar architectural design to conventional FPN, extracting hierarchical feature maps from the backbone network and employing $1\times1$ convolutional operations to reduce their channel dimensions to 64. These channel-reduced features, denoted as \{$X_1$, $X_2$, $X_3$, $X_4$\}, correspond to feature strides of \{2, 4, 8, 16\} pixels with respect to the original input resolution. The multi-scale feature pyramid \{$Y_1$, $Y_2$, $Y_3$, $Y_4$\} is subsequently constructed by the top-down pathway within NS-FPN. Each lateral connection in NS-FPN integrates dual specialized modules for noise suppression: low-frequency guided feature purification (LFP) module and spiral-aware feature sampling (SFS) module. LFP suppresses the noise features by purifying high-frequency components in $X_i$  while SFS takes \{$X'_i$, $Y_{i+1}$\} as input, dynamically samples features in a spiral shape, and calculates similarity to further eliminate the influence of noise interference. Finally, the enhanced features \{$Y_1$, $Y_2$, $Y_3$, $Y_4$\} obtained by feature addition are utilized for the subsequent IRSTDS tasks. Note that, all laterals of NS-FPN contain the LFP module, while only \{$Y_1$, $Y_2$, $Y_3$\} layers contain the SFS module. 

\subsection{Low-frequency Guided Feature Purification}

As observed in the introduction, low-frequency components can be utilized as valuable cues to suppress noise in high-frequency representations. Therefore, we propose the LFP module to reduce false alarms caused by noise interference present in high-frequency features. To achieve this, we design a two-stage purification mechanism, as shown in Figure~\ref{fig:NS-FPN}. In the first stage, we utilize low-frequency features to provide the weighted map of potential target locations, which subsequently guides the enhancement of target-relevant features while suppressing noise features in high-frequency. The second stage refines these high-frequency features through a gated gaussian filtering, which further effectively eliminates noise features. 

% \begin{figure*}[t]
%   \centering
%   \includegraphics[width=\linewidth]{figure/LFP.pdf}
%   \caption{Detailed structure of the low-frequency guided feature purification (LFP) module.}
%   \label{fig:LFP}
% \end{figure*}

Specifically, in the first stage, given the input features $X_i \in \mathbb{R}^{B \times C \times H_i \times W_i}$, we first perform a single-level 2D discrete wavelet transform (DWT) to decompose it into low- and high-frequency features:
\begin{equation}
% [F_l, F_h] = \text{DWT}_{level=1}(X),
[F_l, F_h] = \text{DWT}(X),
\label{eq:dwt}
\end{equation}
where $\mathit{F_l}$ and $\mathit{F_h}$ denote the low- and high-frequency components, respectively. Then we perform a spatial attention to obtain the weighted map from $\mathit{F_l}$ as follows:
\begin{equation}
A_s = \text{Sigmoid}\left( \text{Conv} \left( \text{APool}(F_l) \, \| \, \text{MPool}(F_l) \right) \right),
\label{eq:lf_sa_map}
\end{equation}
where APool and MPool denotes the average and max pooling. The high-frequency features are then modulated via element-wise multiplication: 
\begin{equation}
\hat{F_h} = A_s \odot F_h.
\label{eq:hf_modulation}
\end{equation}

In the second stage, to further purify the modulated high-frequency components $\hat{F_h}$, we introduce a gated gaussian filtering $\mathcal{G}(\cdot)$ to adaptively impose stronger suppression on less confident features. Specifically, we apply gaussian smoothing only to those high-frequency components whose absolute values fall below an empirically set threshold $\tau$:
\begin{equation}
\tilde{F_h} = \mathcal{G}(\hat{F_h}) \cdot \mathbb{I}_{< \tau}(|\hat{F_h}|) + \hat{F_h} \cdot \mathbb{I}_{\geq \tau}(|\hat{F_h}|),
\label{eq:gated_gauss}
\end{equation}
where $\mathbb{I}(\cdot)$ denotes the indicator function used to impose the gating constraint and $\mathcal{G}(\cdot)$ is defined as follows:
\begin{equation}
\mathcal{G}(i, j; \sigma) = \frac{1}{Z} \exp\left( -\frac{(i - c)^2 + (j - c)^2}{2\sigma^2} \right),
\label{eq:gaussian_kernel}
\end{equation}
where $\sigma$ is the learnable standard deviation, $(i, j)$ denotes the spatial coordinate in the kernel, $c = \lfloor k/2 \rfloor$ is the kernel center, and $Z$ is a normalization factor. 
Finally, the purified features are reconstructed by inverse DWT:
\begin{equation}
X' = \text{IDWT}(F_l, \tilde{F_h}).
\label{eq:idwt}
\end{equation}

\subsection{Spiral-aware Feature Sampling}
After purifying each scale feature, we design a spiral-aware feature sampling (SFS) to adaptively acquire the target-relevant features from upper-layer to lower-layer features, thereby further mitigating noise interference in the fusion process. To ensure the consistency of feature scales, feature sampling becomes inevitable. Thus, we integrate the intensity distribution characteristics of IRST and design a spiral pattern to sample target-relevant features in SFS module.

\begin{figure}[!t]
  \centering
  \includegraphics[width=\linewidth]{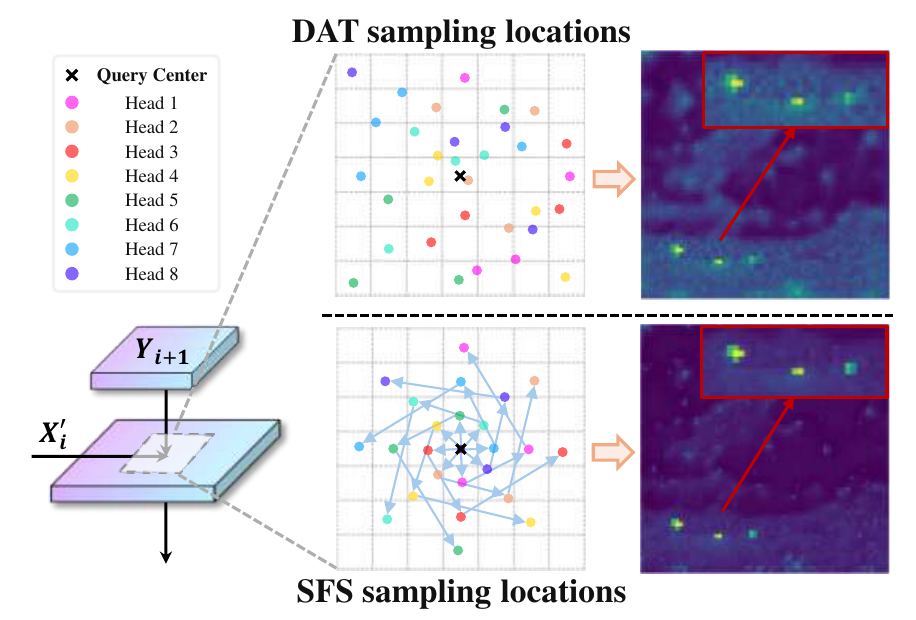} 
  % \caption{Sampling patterns and fusion results of deformable attention (DAT) and spiral-aware fusion sampling (SFS).}
  \caption{Visualization of DAT and SFS sampling process. }
  \label{fig:sampling}
\end{figure}

Specifically, for each LFP-purified feature map $X_i' \in \mathbb{R}^{B \times C \times H_i \times W_i}$, we aim to inject higher-level semantic features $Y_{i+1} \in \mathbb{R}^{B \times C \times H_{i+1} \times W_{i+1}}$ from the deeper layer of NS-FPN. A straightforward way is to use DAT \cite{zhu2020deformable} to randomly sample the $Y_{i+1}$ features to acquire scale-consistent features. However, since IRST is dim and small, random sampling cannot effectively capture the difference between the surrounding area and the target. Thus, we introduce an initialized pattern to constrain the sampling process. We first generate a uniform grid of reference points $p \in \mathbb{R}^{H_G \times W_G \times 2}$ for $Y_{i+1}$, where $H_G \times W_G$ controls the spatial sparsity of the sampling. Then, the offsets $\Delta p$ are generated by combining a basic spiral distribution $s$ and a group of learnable biases $\epsilon$:
\begin{equation}
{Y}_{i+1}'=\phi(Y_{i+1};p+\Delta p),\quad \Delta p = s + \epsilon,
\end{equation}
where $\phi$ denotes the bilinear interpolation, ${Y}_{i+1}'$ is the sampled features. In particular, a spiral pattern is designed to construct $s$, as illustrated in Figure~\ref{fig:sampling}. This pattern is formulated for each attention head $h \in [1, H]$ within the polar coordinate system as follows:
\begin{equation}
\begin{aligned}
s^{(h, k)} =& l_s
\begin{bmatrix} 
\cos(\theta_{h, k}) \ 
\sin(\theta_{h ,k}) 
\end{bmatrix}, \\
\theta_{h, k} = \frac{2\pi k}{P} &+ \frac{2\pi h}{H}, \quad
l_s = l_0 + k \cdot \Delta l,
\end{aligned}
\end{equation}
where $k \in [1, P]$ indexes the sampling point, $l_0$ is the initial radius, and $\Delta l$ is the radial step between consecutive points.

Based on the generated reference points $p$, we can perform multi-head attention to calculate similarity and acquire the fused features. Typically, we utilize the LFP-purified features $X_i'$ as query, and the upper-layer features ${Y}_{i+1}'$ as key and value, which can be represented as:
\begin{equation}
F_{s} = Attn(LN(X_i'),LN(Y_{i+1}')),
\end{equation}
where $Attn(\cdot)$ is the cross attention \cite{carion2020end} and $LN(\cdot)$ is LayerNorm \cite{ba2016layer}. Finally, the output features $Y_i$ are obtained by modulating $X_i'$ with $F_{s}$ through the residual addition:
\begin{equation}
Y_i = X_i' + F_{s}.
\end{equation}

% Unlike \cite{zhu2020deformable} that computes sampling offsets individually for each query, we design a globally shared and structurally initialized pattern to constrain the sampling offsets. 

% feature fusion method 消融
\begin{table}[!t]
\centering
\caption{Comparison of segmentation performance and computation cost between using SFS and other methods.}
\label{tab:fusion_comparison}
\renewcommand{\arraystretch}{1.0}
\begin{tabularx}{\linewidth}{>{\centering\arraybackslash}p{2cm}|>{\centering\arraybackslash}X>{\centering\arraybackslash}X>{\centering\arraybackslash}X|>{\centering\arraybackslash}p{1.3cm}}
\toprule
\textbf{Method} & $IoU\uparrow$ & $P_d\uparrow$ & $F_a\downarrow$ & $Flops(G)$ \\
\midrule
Upsample     & 68.82 & 94.56 & 9.79  & 6.80 \\
% Cross-Attn   &  &  &  & +1.22G\\
DAT \cite{zhu2020deformable}         & 68.52 & 93.54 & 10.40 & +1.24G \\
\textbf{SFS (Ours)} & \underline{\textbf{69.29}} & \underline{\textbf{95.24}} & \underline{\textbf{8.58}} & \underline{\textbf{+1.16G}} \\
\bottomrule
\end{tabularx}
\end{table}

\subsection{Why the SFS Module Works?}
Since using upsampling and pixel-by-pixel addition in the original FPN lacks spatial perception ability around IRST, a straightforward way is to adopt deformable attention (DAT) \cite{zhu2020deformable} to enhance spatial perception while ensuring computational efficiency. However, as shown in Table~\ref{tab:fusion_comparison}, directly using DAT cannot effectively improve performance but increases computational complexity. The main reason is that IRST are typically small, occupy compact and shape-consistent regions, which makes the random sampling in sparse attention not applicable. Therefore, we propose the spiral-aware feature sampling to effectively alleviate the above limitations from two aspects:

\ding{202} \textbf{Spiral spatial perception.}
Since the intensity of IRST has the gaussian distribution characteristics, we explicitly restrict the sampling locations to satisfy the spiral distribution, as illustrated in Figure~\ref{fig:sampling}. This structured-design ensures the network can perceive fine-grained features around the IRST, which yields clearer target contours while cooperating with LFP to further suppress background noise interference as shown in Table~\ref{tab:fusion_comparison}.

\ding{203} \textbf{Shared learnable offsets.}
As analyzed above, IRST are usually characterized by a consistent shape. Thus, different from obvious methods that using learnable offsets individually for each query, SFS module employs a set of shared learnable offsets across different querys. This makes the sampling process more stable while reducing the computational complexity as shown in Table~\ref{tab:fusion_comparison}.

\section{Experiment}
\subsection{Datasets and Evaluation Metrics}
\noindent \textbf{Datasets.}
To evaluate the effectiveness of NS-FPN, we choose IRSTD-1k \cite{zhang2022isnet} and NUAA-SIRST \cite{dai2021asymmetric} as our experimental datasets. IRSTD-1k includes 1000 real infrared images of 512$\times$512 in size, and NUAA-SIRST contains 427 infrared images of various sizes. For each dataset, 80\% of the infrared images are used as training set and the remaining 20\% are used as testing set.

\noindent \textbf{Evaluation Metrics for Segmentation.} For segmentation tasks, Intersection over Union ($IoU$) is used as the pixel-level metric to evaluate shape description capability, and the probability of detection ($P_d$) and the false alarms ($F_a$) rate to evaluate localization performance.

\noindent \textbf{Evaluation Metrics for Detection.} For detection tasks, mean Average Precision ($mAP$) is utilized as the primary metric to evaluate detection performance based on classification accuracy and localization precision. Specifically, $mAP_{50}$ and $mAP_{75}$ denote the $mAP$ at a fixed $IoU$ threshold of 0.50 and 0.75 respectively, while $mAP$ is the average of $mAP$ values calculated over a range of $IoU$ thresholds from 0.50 to 0.95 with a step of 0.05.

\subsection{Implementation Details}
All experiments are conducted on NVIDIA GeForce RTX 4090 GPUs. We integrate NS-FPN into MSHNet \cite{liu2024infrared} for segmentation and YOLOv8n-p2 for detection. The Adagrad optimizer \cite{duchi2011adaptive} is used with an initial learning rate of 0.05. The models are trained for 500 epochs with a batch size of 16. For segmentation tasks, input images are resized to 256 and then randomly cropped to 224 during training. For detection tasks, input images are uniformly resized to 640.
% For loss function, we employ the scale and location sensitive loss (SLSIoULoss) \cite{liu2024infrared} for segmentation and the default YOLOv8 loss configuration for detection. 
% The annotations of the bounding box used for detection are derived from \cite{yang2024eflnet}.

% 模块消融结果
\begin{table}[!t]
\caption{Ablation of LFP and SFS modules with segmentation metrics on IRSTD-1k and NUAA-SIRST datasets.}
\label{tab:ablation_module}
\renewcommand{\arraystretch}{1.0}
\setlength{\tabcolsep}{2mm}
\centering
\begin{tabularx}{\linewidth}{
>{\centering\arraybackslash}X
>{\centering\arraybackslash}X
|>{\centering\arraybackslash}X
>{\centering\arraybackslash}X
>{\centering\arraybackslash}X
|>{\centering\arraybackslash}X
>{\centering\arraybackslash}X
>{\centering\arraybackslash}X}
\toprule
\multicolumn{2}{c|}{\textbf{Module}} & \multicolumn{3}{c|}{\textbf{IRSTD-1k}} & \multicolumn{3}{c}{\textbf{NUAA-SIRST}} \\
\midrule % 分割线
\textbf{LFP} & \textbf{SFS} & $IoU\uparrow$ & $P_d\uparrow$ & $F_a\downarrow$ & $IoU\uparrow$ & $P_d\uparrow$ & $F_a\downarrow$ \\
\midrule
            &             & 67.04 & 91.16 & 13.06  & 76.04 & 99.08 & 12.42 \\
\ding{51}   &             & 68.82 & 94.56 & 9.79  & 76.99 & 100.0 & 12.07 \\
            & \ding{51}   & 67.81 & 93.88 & 13.66  & 78.07 & 100.0 & 4.61  \\
\ding{51}   & \ding{51}   & \underline{\textbf{69.29}} & \underline{\textbf{95.24}} & \underline{\textbf{8.58}} & \underline{\textbf{78.75}} & \underline{\textbf{100.0}} & \underline{\textbf{1.60}} \\
\bottomrule
\end{tabularx}
\end{table}

% 小波增强层数消融结果
\begin{table}[!t]
\centering
\caption{LFP at different scale layers are performed on the IRSTD-1k dataset. The best results are highlighted in \underline{\textbf{bold}}.}
\label{tab:ablation_wav_layers}
\renewcommand{\arraystretch}{1.0}
\setlength{\tabcolsep}{2mm} % 更紧凑的列间距
\begin{tabularx}{\linewidth}{>{\centering\arraybackslash}X
                                 >{\centering\arraybackslash}X
                                 >{\centering\arraybackslash}X
                                 >{\centering\arraybackslash}X
                                 |>{\centering\arraybackslash}X
                                 >{\centering\arraybackslash}X
                                 >{\centering\arraybackslash}X}
\toprule
\multicolumn{4}{c|}{\textbf{LFP at Different Scale Layer}} & \multicolumn{3}{c}{\textbf{IRSTD-1k}} \\
\midrule
\textbf{$X_1$} & \textbf{$X_2$} & \textbf{$X_3$} & \textbf{$X_4$} & $IoU\uparrow$ & $P_d\uparrow$ & $F_a\downarrow$ \\
\midrule
            &             &             &             & 67.81 & 93.88 & 13.66 \\
\ding{51}   &             &             &             & 67.72 & 94.56 & \underline{\textbf{6.15}} \\
\ding{51}   & \ding{51}   &             &             & 67.66 & 94.22 & 8.35 \\
\ding{51}   & \ding{51}   & \ding{51}   &             & 68.45 & 94.22 & 12.98 \\
\ding{51}   & \ding{51}   & \ding{51}   & \ding{51}   & \underline{\textbf{69.29}} & \underline{\textbf{95.24}} & 8.58 \\
\bottomrule
\end{tabularx}
\end{table}

% neck 消融
\begin{table}[!t]
\centering
\caption{Comparison of different FPNs on IRSTD-1k dataset, along with parameters and computational cost.}
\label{tab:neck_comparison}
\renewcommand{\arraystretch}{1.0}
% \begin{tabularx}{\linewidth}{>{\centering\arraybackslash}p{1cm}|*{3}{>{\centering\arraybackslash}X}|*{2}{>{\centering\arraybackslash}X}|*{2}{>{\centering\arraybackslash}X}}
\begin{tabularx}{\linewidth}{
  >{\small\raggedright\arraybackslash}p{1.35cm}   % Method
  |>{\centering\arraybackslash}p{0.4cm}  % IoU
  >{\centering\arraybackslash}p{0.4cm}  % Pd
  >{\centering\arraybackslash}p{0.4cm}  % Fa
  |>{\centering\arraybackslash}p{0.65cm}  % mAP50
  >{\centering\arraybackslash}p{0.65cm}  % mAP
  |>{\centering\arraybackslash}p{0.5cm}  % Param
  >{\centering\arraybackslash}p{0.5cm}  % Flops
}

\toprule
\multirow{2}{*}{\textbf{Method}} & \multicolumn{3}{c|}{\textbf{\small{Segmentation}}} & \multicolumn{2}{c|}{\textbf{\small{Detection}}} & \multicolumn{2}{c}{\textbf{\small{Complexity}}} \\
 & $IoU$ & $P_d$ & $F_a$ & \small{$mAP_{50}$} & \small{$mAP$} & \small{$P(M)$} & \small{$F(G)$} \\

\midrule
FPN\cite{lin2017feature}   & 67.0 & 91.2 & 13.1   & 85.9 & 41.8 & 3.91 & 6.80 \\
PANet\cite{liu2018path} & 68.9 & 93.5 & \underline{\textbf{6.7}} & 85.0 & 41.5 & +0.41 & +1.41 \\
BiFPN\cite{tan2020efficientdet} & 66.9 & 93.5 & 12.1  & 85.8 & 41.6 & +0.39 & +1.33 \\
HSFPN\cite{shi2025hs} & 66.7 & 94.9 & 18.1 & 85.1 & 41.0 & +0.17 & +0.98 \\
\textbf{Ours} & \underline{\textbf{69.2}} & \underline{\textbf{95.2}} & 8.5 & \underline{\textbf{86.3}} & \underline{\textbf{42.1}} & +0.26 & +1.16 \\
\bottomrule
\end{tabularx}
\end{table}

\begin{table*}[!h]
\centering
\caption{Comparison with other state-of-the-art methods on IRSTD-1k and NUAA-SIRST datasets. Segmentation results are evaluated by $IoU$(\%), $P_d$(\%), and $F_a$($10^{-6}$). Detection results are evaluated by $mAP_{50}$(\%), $mAP_{75}$(\%) and $mAP$(\%). The best results are highlighted in \underline{\textbf{bold}} and the second-place results are highlighted in \underline{underline}.}
\label{tab:seg_det_table}
\renewcommand{\arraystretch}{0.85}
\setlength{\tabcolsep}{4pt}

\begin{minipage}{\textwidth}
% ----------------- Segmentation -----------------
\begin{tabularx}{\textwidth}{>{\raggedright\arraybackslash}p{4.5cm}|>{\centering\arraybackslash}p{2.7cm}|>{\centering\arraybackslash}X >{\centering\arraybackslash}X >{\centering\arraybackslash}X|>{\centering\arraybackslash}X >{\centering\arraybackslash}X >{\centering\arraybackslash}X}
\toprule
\multicolumn{8}{c}{\textbf{Segmentation}} \\
\midrule
\multirow{2}{*}{\textbf{Method}} & \multirow{2}{*}{\textbf{Type}} & \multicolumn{3}{c|}{\textbf{IRSTD-1k}} & \multicolumn{3}{c}{\textbf{NUAA-SIRST}} \\
& & $IoU\uparrow$ & $P_d\uparrow$ & $F_a\downarrow$ & $IoU\uparrow$ & $P_d\uparrow$ & $F_a\downarrow$ \\
\midrule
Top-Hat (2010) \cite{bai2010analysis} & \multirow{2}{*}{Filtering} & 10.06 & 75.11 & 1432 & 7.14 & 79.84 & 1012 \\
Max-Median (1999) \cite{deshpande1999max} & & 7.00 & 65.21 & 59.73 & 4.17 & 69.20 & 55.33 \\
\midrule
WSLCM (2020) \cite{han2020infrared} & \multirow{2}{*}{Local Contrast} & 3.45 & 72.44 & 6619 & 1.16 & 77.95 & 5446 \\
TLLCM (2020) \cite{han2020infrared} & & 3.31 & 77.39 & 6738 & 1.03 & 79.09 & 5899 \\
\midrule
IPI (2013) \cite{gao2013infrared} & \multirow{2}{*}{Low Rank} & 27.92 & 81.37 & 16.18 & 25.67 & 85.55 & 11.47 \\
RIPT (2017) \cite{dai2017reweighted} & & 14.11 & 77.55 & 28.31 & 11.05 & 79.08 & 22.61 \\
\midrule
ACMNet (WACV 21) \cite{dai2021asymmetric} & \multirow{12}{*}{Deep Learning} & 59.23 & 93.27 & 65.28 & 70.77 & 93.08 & 3.70 \\
ISNet (CVPR 22) \cite{zhang2022isnet} & & 62.88 & 92.59 & 27.92 & 74.16 & 97.99 & 8.35 \\
DNANet (TIP 22) \cite{li2022dense} & & 65.71 & 91.84 & 17.61 & 74.31 & 98.17 & 15.97 \\
UIUNet (TIP 22) \cite{wu2022uiu} & & 65.06 & 91.16 & 12.68 & 72.69 & \underline{99.08} & 26.61 \\
HCFNet (ICME 24) \cite{xu2024hcf} & & 64.26 & 92.86 & 23.91 & 72.74 & 98.17 & 6.21 \\
IRPruneDet (AAAI 24) \cite{zhang2024irprunedet} & & 64.54 & 91.74 & 16.04 & 75.12 & 98.61 & \underline{2.96} \\
IRSAM (ECCV 24) \cite{zhang2024irsam} & & 64.65 & 90.57 & 16.61 & 71.44 & 92.66 & 7.53 \\
SCTransNet (TGRS 24) \cite{yuan2024sctransnet} & & \underline{68.64} & 91.84 & \underline{11.92} & 77.09 & 98.17 & 15.26 \\
SFCANet (TAES 25) \cite{lin2025sfcanet} & & 66.68 & 92.89 & 12.69 & \underline{78.46} & 97.24 & 8.02 \\
PConv (AAAI 25) \cite{yang2025pinwheel} & & 67.08 & 92.18 & \underline{11.92} & 76.25 & \underline{99.08} & 6.74 \\
MSHNet (CVPR 24) \cite{liu2024infrared} & & 67.16 & \underline{93.88} & 15.03 & 74.60 & \underline{99.08} & 17.21 \\
\textbf{MSHNet + NS-FPN} (Ours) & & \underline{\textbf{69.29}} & \underline{\textbf{95.24}} & \underline{\textbf{8.58}} & \underline{\textbf{78.75}} & \underline{\textbf{100.0}} & \underline{\textbf{1.60}} \\
% \bottomrule % 子表格的底部横线
\specialrule{0pt}{0pt}{0pt}  % 不显示底部横线
\end{tabularx}

% ----------------- Detection -----------------
\begin{tabularx}{\textwidth}{>{\raggedright\arraybackslash}p{4.5cm}|>{\centering\arraybackslash}X >{\centering\arraybackslash}X >{\centering\arraybackslash}X|>{\centering\arraybackslash}X >{\centering\arraybackslash}X >{\centering\arraybackslash}X}
\toprule
\multicolumn{7}{c}{\textbf{Detection}} \\
\midrule
\multirow{2}{*}{\textbf{Method}} & \multicolumn{3}{c|}{\textbf{IRSTD-1k}} & \multicolumn{3}{c}{\textbf{NUAA-SIRST}} \\
       & $mAP_{50}$ & $mAP_{75}$ & $mAP$ & $mAP_{50}$ & $mAP_{75}$ & $mAP$ \\
\midrule
EFLNet (TGRS 24) \cite{yang2024eflnet}       & \underline{86.1} & 29.5 & 38.0 & \underline{\textbf{98.1}} & 37.9 & 47.1 \\
DFLMF-ISTD (IPT 25) \cite{li2025dflmf}      & 82.7 & -- & 38.6 & 94.2 & -- & 51.1 \\
PConv (AAAI 25) \cite{yang2025pinwheel}     & \underline{86.1} & \underline{34.5} & 40.8 & 96.4 & \underline{49.9} & \underline{54.9} \\
YOLOv8n (2024)                   & 85.0 & 31.9 & \underline{41.5} & 95.6 & 40.3 & 49.0 \\
\textbf{YOLOv8n + NS-FPN} (Ours)         & \underline{\textbf{86.3}} & \underline{\textbf{36.9}} & \underline{\textbf{42.1}} & \underline{97.5} & \underline{\textbf{61.6}} & \underline{\textbf{58.0}} \\
\bottomrule
\end{tabularx}
% \vspace{-0.3cm}
\end{minipage}
\end{table*}

% \begin{figure}[t]
%   \centering
%   \includegraphics[width=0.8\linewidth]{figure/gauss_gate.pdf} 
%   \caption{Ablation of different thresholds $\tau$ in LFP module. }
%   \label{fig:gauss_gate}
% \end{figure}

\begin{figure}[t]
  \centering
  \includegraphics[width=\linewidth]{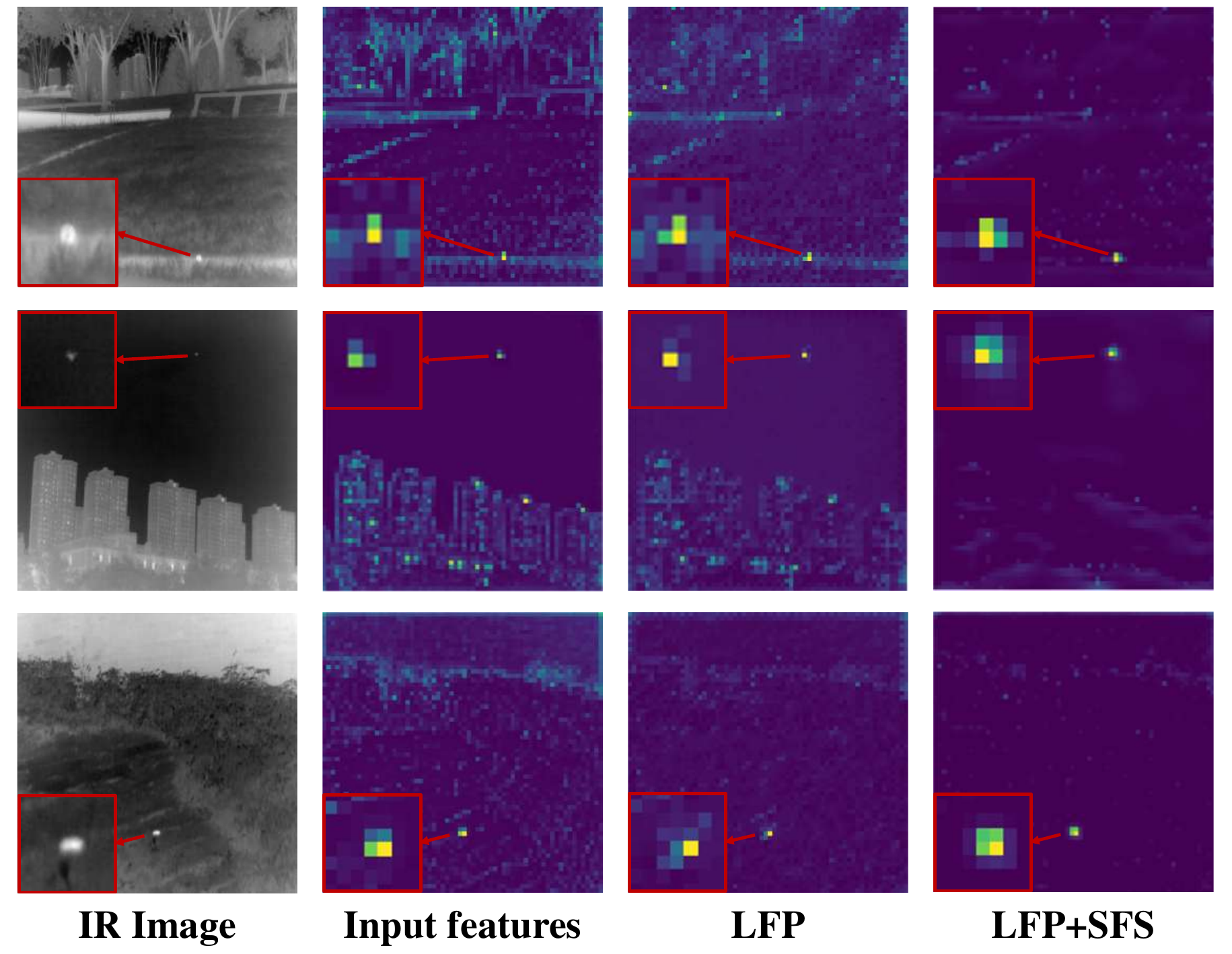} 
  \caption{Visualization of the features at the $X_2$ level after the gradual addition of LFP and SFS in NS-FPN.}
  \label{fig:effect}
\end{figure}

\begin{figure}[t]
  \centering
  \includegraphics[width=1\linewidth]{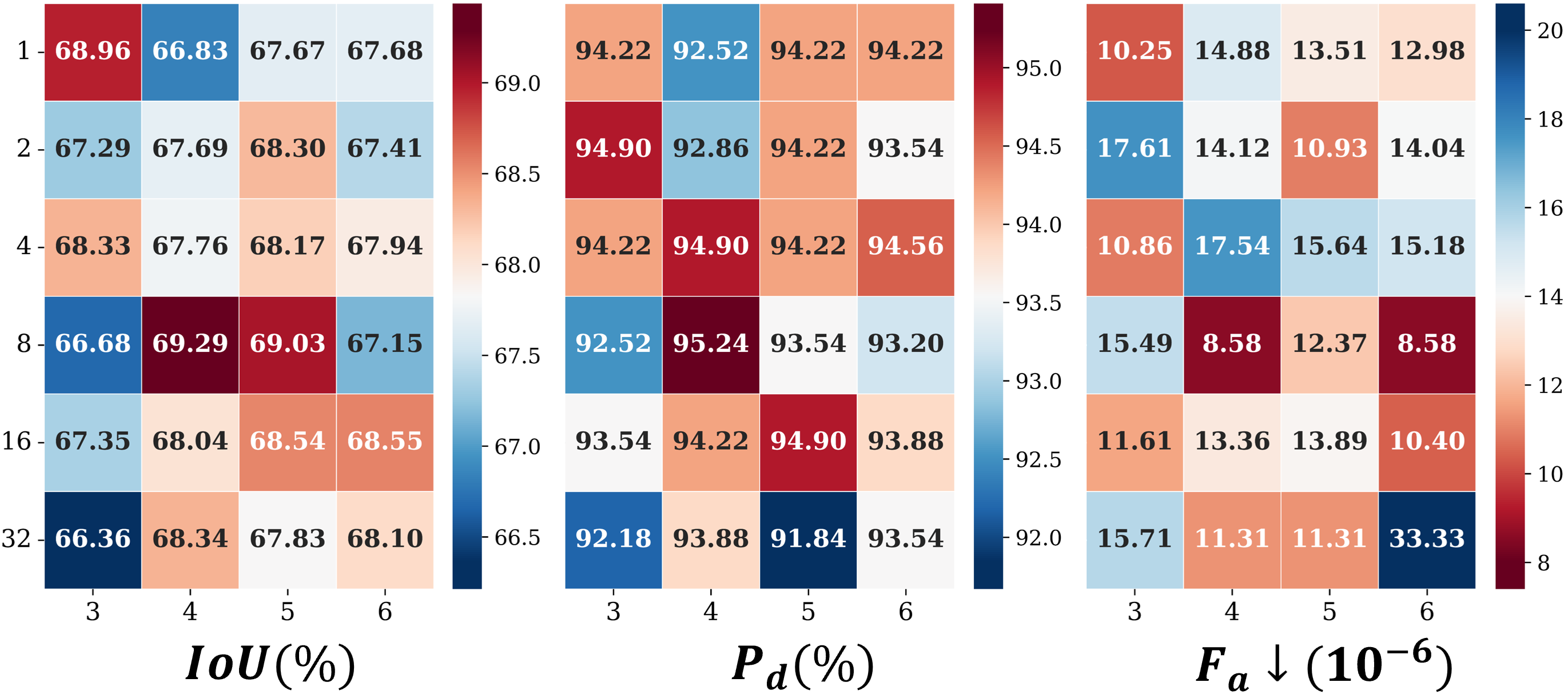} 
  \caption{Heatmaps for the ablation results ($IoU$, $P_d$, $F_a$) of $H$ (Rows) and $P$ (Columns) in SFS module.}
  \label{fig:ablation_hp}
  \vspace{-0.3cm}
\end{figure}

\subsection{Ablation Study}

\noindent \textbf{Ablation on Each Component.}
To evaluate the effectiveness of the NS-FPN, we conduct ablation on the LFP and SFS modules. The baseline model is MSHNet and YOLOv8n-p2 with FPN and the results are shown in Table~\ref{tab:ablation_module}. Applying LFP or SFS to baseline model individually improves the segmentation performance. Specifically, replacing the 1×1 convolution in FPN with the LFP module significantly improves all metrics. In particular, on IRSTD-1k, $IoU$ increases by 1.78\%, $P_d$ increases by 3.40\%, and $F_a$ decreases by 3.27. The best results are achieved when using both LFP and SFS modules, yielding a significant improvement in $F_a$ on both datasets. As shown in Figure~\ref{fig:effect}, the combination of LFP and SFS effectively suppress the background noise and enhance the target features.
% LFP enhances the target response while effectively suppressing background clutter. Upon this, SFS injects higher-level semantic features to further reduce false alarms and highlight the target more distinctly. These improvements demonstrate the effectiveness of proposed LFP and SFS modules.

% These results demonstrate that LFP effectively exploits low-frequency components to guide the selection of informative high-frequency features, helping the model to focus more precisely on true targets while suppressing background interference.
% When both LFP and SFS are applied, we observe the best overall performance. This confirms the significant impact of the complete NS-FPN architecture, which jointly enhances the model’s capability in both pixel-wise segmentation and object-level detection.

\noindent \textbf{LFP at Different Scale Layer.}
We use the proposed LFP to replace the 1$\times$1 convolutions of different scale layers in FPN. Table~\ref{tab:ablation_wav_layers} presents the results of applying LFP module at different layers (from $X_1$ to $X_4$). We observe that applying LFP to large-scale layers ($X_1$, $X_2$) significantly reduces $F_a$ to 6.15. In contrast, small-scale layers ($X_3$, $X_4$) capture more semantic context, improving $IoU$ but with higher $F_a$. When LFP is applied to all scale layers, the best overall trade-off of segmentation metrics can be achieved.

\noindent \textbf{Ablation of the hyperparameters $H$ and $P$ in SFS.}
To determine the optimal hyperparameter group of $H$ and $P$ in our SFS modules, we perform ablation studies on IRSTD-1k dataset by varying the number of heads $H \in \{1, 2, 4, 8, 16, 32\}$ and points $P \in \{3, 4, 5, 6\}$, where the FPN channel dimensions of $64$ should be divisible by $H$. As detailed in Figure~\ref{fig:ablation_hp}, increasing $H$ moderately improves performance, but excessive $H$ leads to a performance drop due to insufficient information per head. Similarly, moderate increases in $P$ improve $P_d$, while an overly large $P$ introduces more computation and false alarms. Finally, we select the optimal configuration of $H=8$ and $P=4$ for SFS module, where all metrics achieve the best performance.

\noindent \textbf{Ablation on Different FPNs.}
To demonstrate the superiority of our NS-FPN, we compare it with representative alternatives including FPN \cite{lin2017feature}, PANet \cite{liu2018path}, BiFPN \cite{tan2020efficientdet} and HS-FPN \cite{shi2025hs}, as shown in Table~\ref{tab:neck_comparison}. For segmentation, our method achieves the highest $IoU$ (69.29\%) and $P_d$ (95.24\%), while maintaining acceptable $F_a$. For detection, NS-FPN also achieves the best $mAP_{50}$ (86.3\%) and $mAP$ (42.1\%). These results demonstrate that NS-FPN is an effective plugin designed specifically for the IRSTDS tasks.
% The results confirm our NS-FPN not only enhances multi-scale feature fusion tailored to the characteristics of infrared small targets, but also serves as a general, plug-and-play neck with superior performance in both segmentation and detection tasks across different infrared scenarios.

\begin{figure*}[t]
  \centering
  \includegraphics[width=1\textwidth]{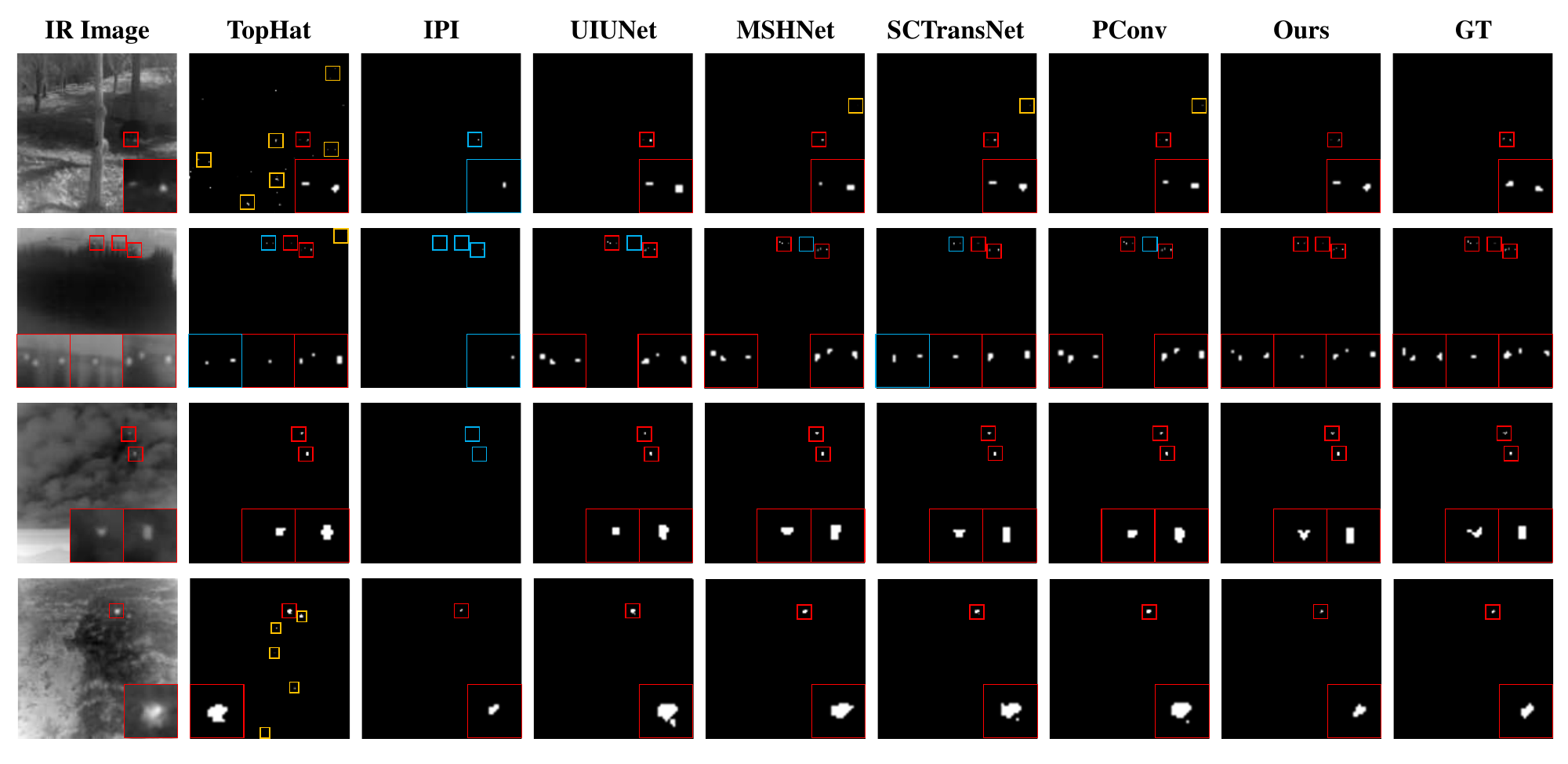}
  \caption{Visual results of different SOTA methods. The boxes in \textbf{\textcolor[RGB]{255,0,0}{red}}, \textbf{\textcolor[RGB]{0,176,240}{blue}}, and \textbf{\textcolor[RGB]{255,192,0}{yellow}} represent correct, missed, and false detections, respectively. Close-up views are shown in the bottom corners.}
  \label{fig:vis_comparison}
\end{figure*}

\subsection{Comparison with SOTA Methods}

\noindent \textbf{Quantitative Comparisons.}
% \noindent \textbf{Segmentation:}
For segmentation tasks, we select six classical methods and eleven state-of-the-art deep learning methods, including several frequency domain-based models, for comparisons on the IRSTD-1k and NUAA-SIRST datasets. As presented in Table~\ref{tab:seg_det_table}, traditional methods such as Top-Hat \cite{bai2010analysis} and WSLCM \cite{han2020infrared} have limited performance due to their reliance on handcrafted priors. For deep learning-based methods, our method achieves the best performance in all evaluation metrics on both datasets, with 69.29\% $IoU$, 95.24\% $P_d$, and 8.58 $F_a$ on IRSTD-1k, and 78.75\% $IoU$, 100.0\% $P_d$, and 1.60 $F_a$ on NUAA-SIRST, demonstrating superior performance in pixel-level accuracy and object-level reliability. 
As for detection tasks, we further compare with four recent IRSTD methods on IRSTD-1k and NUAA-SIRST datasets. Table~\ref{tab:seg_det_table} shows that our approach outperforms previous methods in $mAP_{50}$, $mAP_{75}$ and $mAP$. Specifically, our method achieves the best results on both datasets with 86.3\% $mAP_{50}$, 36.9\% $mAP_{75}$ and 42.1\% $mAP$ on IRSTD-1k, and 97.5\% $mAP_{50}$, 61.6\% $mAP_{75}$ and 58.0\% $mAP$ on NUAA-SIRST. The above results demonstrate that our NS-FPN is highly effective for small target segmentation and detection tasks in infrared scenarios, enabling accurate target localization while suppressing background noise interference.

% 模块消融结果
\begin{table}[!t]
\caption{Segmentation performance, FLOPs and parameters comparison for NS-FPN components.}
\label{tab:module_cost}
\renewcommand{\arraystretch}{1.0}
\setlength{\tabcolsep}{2mm}
\centering
\begin{tabularx}{\linewidth}{
>{\centering\arraybackslash}p{0.6cm}
>{\centering\arraybackslash}p{0.6cm}
|*{3}{>{\centering\arraybackslash}p{0.7cm}}|*{2}{>{\centering\arraybackslash}X}}
\toprule
\textbf{LFP} & \textbf{SFS} & $IoU\uparrow$ & $P_d\uparrow$ & $F_a\downarrow$ & $Param$ & $FLOPs$ \\
\midrule
            &             & 67.04 & 91.16 & 13.06  & 3.91M & 6.80G \\
\ding{51}   &             & 68.82 & 94.56 & 9.79  & +0.01M & +0.01G \\
            & \ding{51}   & 67.81 & 93.88 & 13.66  & +0.25M & +1.15G \\
\ding{51}   & \ding{51}   & \underline{\textbf{69.29}} & \underline{\textbf{95.24}} & \underline{\textbf{8.58}} & +0.26M & +1.16G \\
\bottomrule
\end{tabularx}
\vspace{-0.3cm}
\end{table}

\noindent \textbf{Visual Comparisons.}
Visual results with closed-up views of different methods are shown in Figure~\ref{fig:vis_comparison}. By using NS-FPN, our method significantly reduces false alarms (the first line), achieves more accurate segmentation results of the target shape (the third line) and effectively distinguishes targets from complex background interference (the fourth line) without missing the dim target (the second line). In contrast, traditional methods often suffer from numerous false alarms and missed detections, while other deep learning-based approaches tend to produce ambiguous predictions. Results above highlight the effectiveness of our proposed NS-FPN in enhancing discriminative features and suppressing interference, enabling precise segmentation even under complex background interference.

% \noindent \textbf{Detection:}
% We further compare our method with four recent IRSTD approaches on IRSTD-1k and NUAA-SIRST, to verify the generalization of our feature fusion strategy. \cref{tab:seg_det_table} shows that our approach outperforms previous methods in both $mAP_{50}$ and $mAP$. Specifically, our method achieves the best results on both datasets with 86.3\% $mAP_{50}$ and 42.1\% $mAP$ on IRSTD-1k, and 97.1\% $mAP_{50}$ and 57.2\% $mAP$ on NUAA-SIRST.
% %%%%%%%%%%%%%%%%%%%%%%%%%%%%%%%%%%
% Compared to YOLOv8n-p2, our method improves $mAP$ on IRSTD-1k by +1.3\% and on NUAA-SIRST by +15.4\%, while maintaining a compact model size of 2.78M and moderate computation of 13.4G $FLOPs$. This demonstrates the strong generalization ability and robustness of NS-FPN under varying infrared imaging conditions.

\subsection{Model Complexity Analysis}
Our paper aims to suppress noise interference in feature fusion process by introducing a lightweight yet effective FPN. As shown in Table~\ref{tab:neck_comparison}, we can see that NS-FPN has superior segmentation and detection performance compared to other FPNs while ensuring low computational complexity, which verifies the efficiency of NS-FPN. 
Furthermore, we also analyze the parameters and computational complexity of individual NS-FPN component as shown in Table~\ref{tab:module_cost}. The parameters and FLOPs for LFP module mainly come from the few convolution operations, which results in similar computational costs with the original FPN. Although the SFS module adopts a cross-attention mechanism that increases FLOPs, the overall NS-FPN framework achieves significant performance improvements at an acceptable computational cost compared to the original FPN.

% \subsection{Visualization of Intermediate Results(option)}
% In \cref{fig:vis_comparison}, we present some infrared small target segmentation results from different methods. Our architecture achieves more accurate segmentation of the target shape (the first and third images) and effectively distinguishes true targets from complex background interference (the fourth image) without ignoring a dim target(the second image). In contrast, traditional methods often suffer from numerous false alarms and missed detections, while other deep learning-based approaches tend to produce ambiguous predictions. These results highlight the effectiveness of our proposed NS-FPN in enhancing discriminative features and suppressing interference, enabling precise segmentation even under challenging conditions.

\section{Conclusion}
In this paper, we proposed a noise-suppression feature pyramid network (NS-FPN) to improve the IRSTDS performance. Given the noise interference results in false alarms problems, we introduced a low-frequency guided feature purification module (LFP) to suppress the noise features by purifying high-frequency components. Additionally, we designed a spiral-aware feature sampling module (SFS) to spirally sample multi-scale features for feature fusion. These two modules are tightly coupled in the FPN framework to improve detection and segmentation performance. Extensive experiments showed that our method surpasses SOTA methods in objective metrics and subjective evaluations. We believe that NS-FPN paves the way for more robust and practical IRSTDS methods in real-world applications.

{
    \small
    \bibliographystyle{ieeenat_fullname}
    \bibliography{main}
}


\end{document}